\documentclass[11pt,a4paper]{article}
\usepackage[hyperref]{acl2019}
\usepackage{times}
\usepackage{latexsym}
\usepackage{booktabs}
\usepackage{graphicx}
\usepackage{amssymb}
\usepackage{amsmath}
\usepackage{pifont}
\usepackage{url}
\usepackage[symbol]{footmisc}
\usepackage{bm}
\usepackage{multirow}
\usepackage{pstool}

\newcommand{\commentout}[1]{}

\usepackage{url}

\aclfinalcopy 

\title{Synthetic QA Corpora Generation with Roundtrip Consistency}

\author{Chris Alberti
\;\;\;\;
  Daniel Andor 
\;\;\;\;
  Emily Pitler 
\;\;\;\;
  Jacob Devlin 
\;\;\;\;
  Michael Collins \\
Google Research\\
{ \small   \texttt{\{chrisalberti, andor, epitler, jacobdevlin, mjcollins\}@google.com}}}
\date{}

\begin{document}
\maketitle
\begin{abstract}
We introduce a novel method of generating synthetic question answering corpora by combining models of question generation and answer extraction, and by filtering the results to ensure roundtrip consistency. By pretraining on the resulting corpora we obtain significant improvements on SQuAD2 \cite{rajpurkar2018know} and NQ \cite{kwiatkowski2019nq}, establishing a new state-of-the-art on the latter. Our synthetic data generation models, for both question generation and answer extraction, can be fully reproduced by finetuning a publicly available BERT model \cite{devlin2018bert} on the extractive subsets of SQuAD2 and NQ.
We also describe a more powerful variant that does full sequence-to-sequence pretraining for question generation, obtaining exact match and F1 at less than $0.1\%$ and $0.4\%$ from human performance on SQuAD2.
\end{abstract}

\section{Introduction}

Significant advances in Question Answering (QA) have recently been achieved by pretraining deep transformer language models on large amounts of unlabeled text data, and finetuning the pretrained models on hand labeled QA datasets, e.g.\ with BERT \cite{devlin2018bert}.

Language modeling is however just one example of how an auxiliary prediction task can be constructed from widely available natural text, namely by masking some words from each passage and training the model to predict them. It seems plausible that other auxiliary tasks might exist that are better suited for QA, but can still be constructed from widely available natural text. It also seems intuitive that such auxiliary tasks will be more helpful the closer they are to the particular QA task we are attempting to solve.

\begin{table}[t]
    \centering
    \resizebox{\columnwidth}{!}{%
    \begin{tabular}{ll}
    \toprule
    \multirow{3}{*}{\bf Input (C)} & ... in 1903, boston participated in the \\ 
    & first modern
    world series, going up  \\  
     &  against the pittsburgh pirates ...   \\
  (1) \quad {\bf $\bm{C \rightarrow A}$} & 1903 \\
(2) \quad   {\bf $\bm{C, A \rightarrow Q}$} & when did the red sox first go to the \\
    & world series \\
  (3) \quad {\bf $\bm{C, Q \rightarrow A'}$}&  1903 \\
 (4) \quad  {\bf $\bm{A \stackrel{?}{=} A'}$} & Yes \\
    \bottomrule
    \end{tabular}}
    \caption{Example of how synthetic question-answer pairs are generated.  The model's predicted answer ($A'$) matches the original answer the question was generated from, so the example is kept.}
    \label{tab:example}
\end{table}

Based on this intuition we construct  auxiliary tasks for QA, generating millions of synthetic question-answer-context triples from unlabeled passages of text, pretraining a model on these examples, and finally finetuning on a particular labeled dataset. Our auxiliary tasks are illustrated in Table 
\ref{tab:example}.

For a given passage $C$, we sample an extractive short answer $A$ (Step (1) in Table \ref{tab:example}).
In Step (2), we generate a question $Q$
conditioned on $A$ and $C$, then (Step (3)) predict the extractive answer $A'$ conditioned on $Q$ and $C$. 
If $A$ and  $A'$ match we finally emit $(C, Q, A)$ as a new synthetic training example (Step (4)). 
We train a separate model on labeled QA data for each of the first three steps, and then apply the models in sequence on a large number of unlabeled text passages. 
We show that pretraining on synthetic data generated through this procedure provides us with significant improvements on two challenging datasets, SQuAD2 \cite{rajpurkar2018know} and NQ \cite{kwiatkowski2019nq}, achieving a new state of the art on the latter.

\section{Related Work}
Question generation is a well-studied task in its own right \cite{heilman2010good, du2017, du2018harvesting}.  
\newcite{yang2017semi} and \newcite{dhingra2018simple} both use generated question-answer pairs to improve a QA system, showing large improvements in low-resource settings with few gold labeled examples.
Validating and improving the accuracy of these generated QA pairs, however, is relatively unexplored.

In machine translation, modeling consistency with dual learning \cite{he2016dual} or back-translation \cite{sennrich2016improving} across both translation directions improves
the quality of translation models.
Back-translation, which adds synthetically
generated parallel data as training examples, was an inspiration for this work,
and has led to state-of-the-art results in  both
the supervised \cite{edunov2018} and the unsupervised settings \cite{lample2018phrase}.

\newcite{lewis2018generative} model the joint distribution of questions and answers given a context and use this model directly, whereas our work uses generative models to generate synthetic data to be used for pretraining. Combining these two approaches could be an area of fruitful future work.

\newcommand{\CtoA}{\theta_A}   
\newcommand{\CAtoQ}{\theta_Q}  
\newcommand{\CQtoA}{\theta_{A'}}   
\newcommand{\eCtoA}{\hat{\theta}_A}   
\newcommand{\eCAtoQ}{\hat{\theta}_Q}  
\newcommand{\eCQtoA}{\hat{\theta}_{A'}}   
\newcommand{\aux}{\beta}

\section{Model}


Given a dataset of contexts, questions, and answers: $\{(c^{(i)}, q^{(i)}, a^{(i)}): i = 1, \ldots, N\}$,
we train
three models: (1) answer extraction: $p(a | c; \CtoA)$, 
    (2) question generation: $p(q | c, a; \CAtoQ)$, and
    (3) question answering: $p(a | c, q; \CQtoA)$.

We use BERT \cite{devlin2018bert}\footnote{Some experiments use a variant of BERT that masks out whole words at training time, similar to \newcite{Sun2019ERNIEER}.  See \url{https://github.com/google-research/bert} for both the original and whole word masked versions of BERT.} to model each of these distributions. 
Inputs to each of these models are fixed length sequences of wordpieces, listing the tokenized question (if one was available) followed by the context $c$. 
The answer extraction model is detailed in \S\ref{extractive-qa} and two variants of question generation models in \S\ref{encoder-only}
and \S\ref{seq2seq}.
The question answering model follows \newcite{alberti2019bert}.

\subsection{Question (Un)Conditional Extractive QA}
\label{extractive-qa}


We define a question-unconditional extractive answer model $p(a|c;\CtoA)$ and a question-conditional extractive answer model $p(a|q,c;\CQtoA)$ as follows:
\begin{align*}
    p(a|c;\CtoA) &=
    \frac{e^{f_J(a, c; \CtoA)}}
    {\sum_{a''} e^{f_J(a'', c; \CtoA)}}
    \\
    p(a|c,q;\CQtoA) &=
    \frac{e^{f_I(a, c, q; \CQtoA)}}
    {\sum_{a''} e^{f_I(a'', c, q; \CQtoA)}}
\end{align*}
where $a$, $a''$ are defined to be token spans over $c$. For $p(a | c;\CtoA)$, $a$ and $a''$ are constrained to be of length up to $L_A$, set to 32 word piece tokens. The key difference between the two expressions is that $f_I$ scores the start and the end of each span independently, while $f_J$ scores them jointly.

Specifically we define $f_J: \mathbb{R}^h \rightarrow \mathbb{R}$ and $f_I: \mathbb{R}^h \rightarrow \mathbb{R}$ to be transformations of the final token representations computed by a BERT model:
\begin{align*}
    & f_J (a, c; \CtoA) = \\
    & \;\; \mbox{MLP}_J(\mbox{CONCAT}(\mbox{BERT}(c)[s], \mbox{BERT}(c)[e]))\\
    & f_I (a, q, c; \CQtoA)) = \\
    & \;\; \mbox{AFF}_I(\mbox{BERT}(q, c)[s]) + \mbox{AFF}_I(\mbox{BERT}(q, c)[e]).
\end{align*}
Here $h$ is the hidden representation dimension, $(s, e) = a$ is the answer span, $\mbox{BERT}(t)[i]$ is the BERT representation of the $i$'th token in token sequence $t$. ${\rm MLP}_J$ is a multi-layer perceptron with a single hidden layer, and ${\rm AFF}_I$ is an affine transformation.

We found it was critical to model span start and end points jointly in $p(a|c;\CtoA)$ because, when the question is not given, there are usually multiple acceptable answers for a given context, so that the start point of an answer span cannot be determined separately from the end point.


\subsection{Question Generation: Fine-tuning Only} \label{encoder-only}


Text generation allows for a variety of choices in model architecture and training data. In this section we opt for a simple adaptation of the public BERT model for text generation. This adaptation does not require any
additional pretraining and no extra parameters need to be trained from scratch at finetuning time. 
This question generation system can be 
reproduced by simply finetuning a publicly available pretrained BERT model on the extractive subsets of datasets like SQuAD2 and NQ.

\paragraph{Fine-tuning} We define the $p(q | c, a; \CAtoQ)$ model as a left-to-right language model
\begin{align*}
    p(q|a, c; \CAtoQ) & = \prod_{i=1}^{L_Q} p(q_i | q_1, \ldots, q_{i-1}, a, c; \CAtoQ) \\
    & = \prod_{i=1}^{L_Q}
    \frac{e^{f_Q(q_1, \ldots, q_i, a, c; \CAtoQ)}}{\sum_{q_i'} e^{f_Q(q_1, \ldots, q_i', a, c; \CAtoQ)}},
\end{align*}
where $q = (q_1, \ldots, q_{L_Q})$ is the sequence of question tokens and $L_Q$ is a predetermined maximum question length, but, unlike the more usual encoder-decoder approach, we compute $f_Q$ using the single encoder stack from the BERT model:
\begin{align*}
  f_Q & (q_1, \ldots, q_i, a, c; \CAtoQ) = \\
  &{\rm BERT}(q_1, \ldots, q_{i-1}, a, c)[i-1] \cdot W_{{\rm BERT}}^\intercal,
\end{align*}
where $W_{{\rm BERT}}$ is the word piece embedding matrix in BERT. All parameters of BERT including $W_{{\rm BERT}}$ are finetuned. In the context of question generation, the input answer is encoded by introducing a new token type id for the tokens in the extractive answer span, e.g. the question tokens being generated have type 0 and the context tokens have type 1, except for the ones in the answer span that have type 2. We always pad or truncate the question being input to BERT to a constant length $L_Q$ to avoid giving the model information about the length of the question we want it to generate.

This model can be trained efficiently by using an attention mask that forces to zero all the attention weights from $c$ to $q$ and from $q_i$ to $q_{i+1} \ldots q_{L_Q}$ for all $i$. 

\paragraph{Question Generation} At inference time we generate questions through iterative greedy decoding, by computing $\operatorname{argmax}_{q_i}\,f_Q (q_1, \ldots, q_i, a, c)$ for $i = 1, \ldots, L_Q$.
Question-answer pairs are kept only if they satisfy roundtrip consistency.

\subsection{Question Generation: Full Pretraining}
\label{seq2seq}
The prior section addressed a restricted setting in which a BERT model was fine-tuned, without any further changes.  
In this section, we describe an alternative approach for question generation that fully pretrains and fine-tunes a sequence-to-sequence generation model.

\paragraph{Pretraining} Section \ref{encoder-only} used only an encoder for question generation.  In this section, we use a full sequence-to-sequence Transformer (both encoder and decoder).  
The encoder is trained identically (BERT pretraining, Wikipedia data), while the decoder is trained to output the next sentence.

\paragraph{Fine-tuning} Fine-tuning is done identically as in Section \ref{encoder-only}, where the input is $(C, A)$ and the output is $Q$ from tuples from a supervised question-answering dataset (e.g., SQuAD).

\paragraph{Question Generation} To get examples of synthetic $(C, Q, A)$ triples,
we sample from the decoder with both beam search and Monte Carlo search.
As before, we use roundtrip consistency to keep only the high precision triples.


\subsection{Why Does Roundtrip Consistency Work?} \label{roundtrip}

A key question for future work is to develop a more formal understanding
of why the roundtrip method improves accuracy on question answering tasks
(similar questions arise for the back-translation methods of \newcite{edunov2018} and \newcite{sennrich2016improving}; 
a similar theory may apply to these methods). In the supplementary material we sketch
a possible approach, inspired by the method of \newcite{Balcan:2005} for 
learning with labeled and unlabeled data. This section is intentionally 
rather speculative but is intended to develop intuition about the methods, and
to propose possible directions for future work on developing a formal grounding.

In brief, the approach discussed in the supplementary material suggests 
optimizing the log-likelihood of the labeled training examples, under 
a constraint that some measure of roundtrip consistency $\aux(\CQtoA)$ on unlabeled
data is greater than some value $\gamma$. The value for $\gamma$ can be estimated
using performance on development data. The auxiliary function $\aux(\CQtoA)$
is chosen such that: (1) the constraint $\aux(\CQtoA) \geq \gamma$ eliminates
a substantial part of the parameter space, and hence reduces sample complexity;
(2) the constraint $\aux(\CQtoA) \geq \gamma$ nevertheless includes `good' parameter
values that fit the training data well. The final step
in the argument is to make the case that the algorithms described in the current
paper may effectively be optimizing a criterion of this kind. Specifically,
the auxiliary function $\aux(\CQtoA)$ is defined as the log-likelihood of
noisy $(c, q, a)$ triples generated from unlabeled data using the
$C \rightarrow A$ and $C, A \rightarrow Q$ models; constraining the parameters
$\CQtoA$ to achieve a relatively high value on $\aux(\CQtoA)$ is achieved
by pre-training the model on these examples. Future work should
consider this connection in more detail.

\section{Experiments} \label{experiments}

\subsection{Experimental Setup}


\begin{table}
\begin{center}
\resizebox{\columnwidth}{!}{%
 \begin{tabular}{l c c c c} 
 \toprule
   & \multicolumn{2}{c}{Dev}
   & \multicolumn{2}{c}{Test} \\
   & EM & F1 & EM & F1 \\
 \midrule
   \it Fine-tuning Only \\
   BERT-Large (Original) & 78.7 & 81.9 & 80.0 & 83.1 \\
 \ \   + 3M synth SQuAD2 & 80.1 & 82.8 & - & - \\
 \ \ \ \   + 4M synth NQ & 81.2 & 84.0 & 82.0 & 84.8 \\
   \it Full Pretraining  \\
   BERT (Whole Word Masking)\footnotemark & 82.6 & 85.2 & - & - \\
 \ \   + 50M synth SQuAD2 & 85.1 & 87.9 & 85.2 & 87.7\\
   \ \ \ \  + ensemble & 86.0 & 88.6 & 86.7 & 89.1\\ 
 \midrule
   Human & - & - & 86.8 & 89.5 \\
 \bottomrule
\end{tabular}}
\end{center}
\caption{Our results on SQuAD2. For our fine-tuning only setting, we compare a BERT baseline (BERT single model - Google AI Language on the SQuAD2 leaderboard) 
to similar models pretrained on our synthetic SQuAD2-style corpus and on a corpus containing both SQuAD2- and NQ-style data. For the full pretraining setting, we report our best single model and ensemble results.}
\label{tab:squad2_results}
\end{table}

\begin{figure}[t]
\centering
\includegraphics[width=0.5\textwidth]{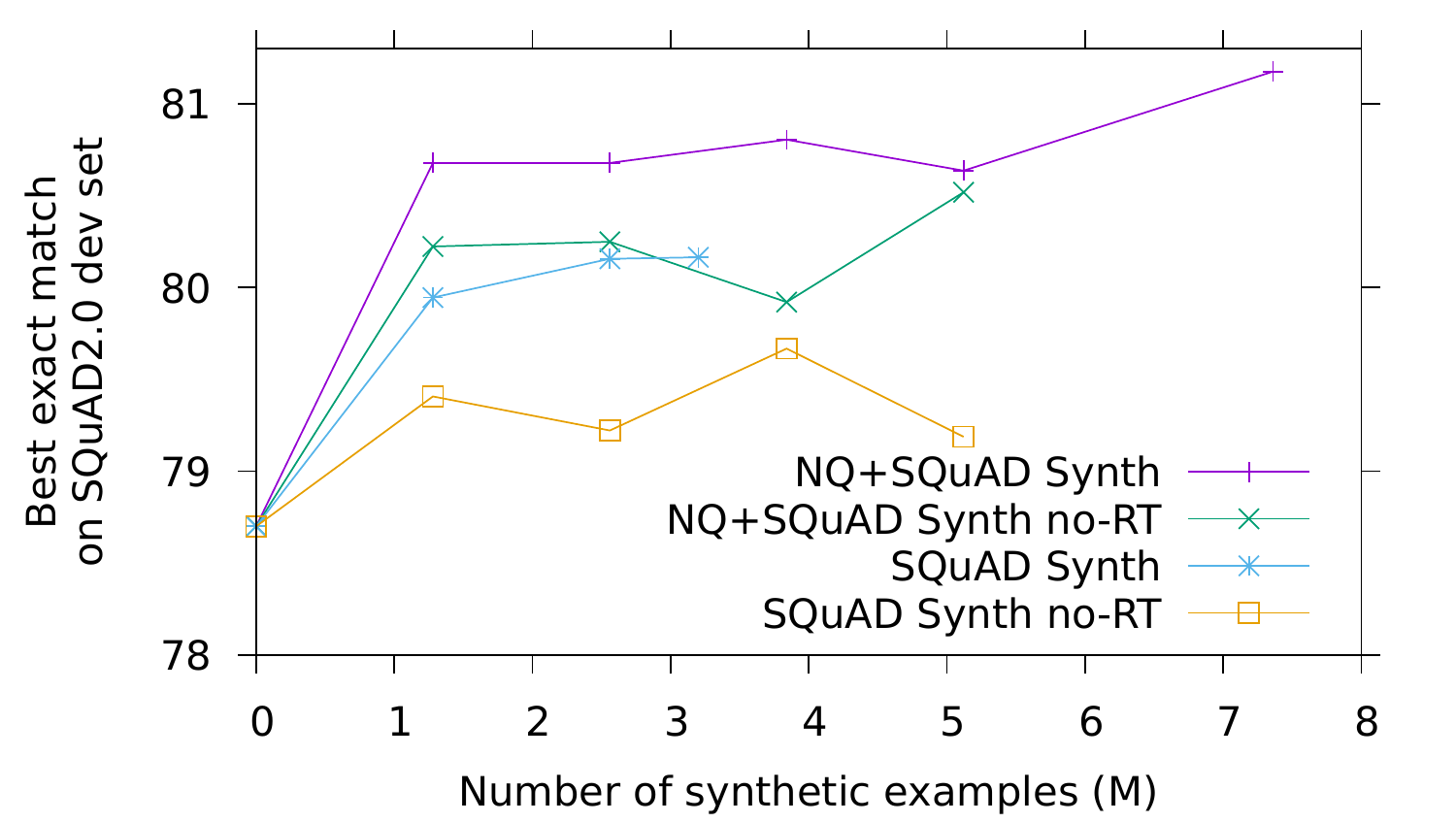}
\caption{Learning curves for pretraining using synthetic question-answering data (fine-tuning only setting). ``no-RT'' refers to omitting the roundtrip consistency check. Best exact match is reported after fine-tuning on SQuAD2. Performance improves with the amount of synthetic data.
For a fixed amount of synthetic data,
having a more diverse source (NQ+SQuAD vs. just SQuAD) yields higher accuracies.
Roundtrip filtering gives further improvements.
}
\label{fig:learning-curves}
\end{figure}

\begin{table*}
\begin{center}
\resizebox{\textwidth}{!}{%
 \begin{tabular}{l c c c c c c c c c c c c c} 
 \toprule
   & \multicolumn{3}{c}{Long Answer Dev}
   & \multicolumn{3}{c}{Long Answer Test} &
   & \multicolumn{3}{c}{Short Answer Dev}
   & \multicolumn{3}{c}{Short Answer Test} \\
   &    P &    R &   F1 &    P &    R &   F1 &
   &    P &    R &   F1 &    P &    R &   F1 \\
 \midrule
   BERT\textsubscript{joint}
   & 61.3 & 68.4 & 64.7 & 64.1 & 68.3 & 66.2 &
   & 59.5 & 47.3 & 52.7 & \textbf{63.8} & 44.0 & 52.1 \\
   + 4M synth NQ & \textbf{62.3} & \textbf{70.0} & \textbf{65.9} &
   \textbf{65.2}  & \textbf{68.4} & \textbf{66.8} & & \textbf{60.7} & \textbf{50.4} & \textbf{55.1} &
    62.1 & \textbf{47.7} & \textbf{53.9}  \\
 \midrule
   Single Human
   & 80.4 & 67.6 & 73.4 & -    & -    & -    &
   & 63.4 & 52.6 & 57.5 & -    & -    & -    \\
   Super-annotator
   & 90.0 & 84.6 & 87.2 & -    & -    & -    &
   & 79.1 & 72.6 & 75.7 & -    & -    & -    \\
 \bottomrule
\end{tabular}}
\end{center}
\caption{Our results on NQ, compared to the previous best system and to
the performance of a human annotator and of an ensemble of human annotators. BERT\textsubscript{joint} is the model described in \newcite{alberti2019bert}.}
\label{tab:nq_results}
\end{table*}


\begin{table*}
\begin{center}
\resizebox{\textwidth}{!}{%
 \begin{tabular}{p{2cm} p{8cm} p{5cm}} 
 \toprule
   & Question & Answer \\
  \midrule
   NQ &
   what was the population of chicago in 1857? &
   over 90,000 \\
   SQuAD2  &
   what was the weight of the brigg's hotel? &
   22,000 tons \\
  \midrule
  NQ & where is the death of the virgin located? & louvre \\
  SQuAD2 & what person replaced the painting? & carlo saraceni \\
  \midrule
  NQ & when did rick and morty get released? & 2012 \\
  SQuAD2 & what executive suggested that rick be a grandfather? & nick weidenfeld \\
 \bottomrule
\end{tabular}}
\end{center}
\caption{Comparison of question-answer pairs generated by NQ and SQuAD2 models for the same passage of text.}
\label{tab:samples}
\end{table*}

We considered two datasets in this work: SQuAD2 \cite{rajpurkar2018know} and the Natural Questions (NQ) \cite{kwiatkowski2019nq}. SQuAD2 is a dataset of QA examples of questions with answers formulated and answered by human annotators about Wikipedia passages. NQ is a dataset of Google queries with answers from Wikipedia pages provided by human annotators. We used the full text from the training set of NQ (1B words) as a source of unlabeled data.

In our fine-tuning only experiments (Section \ref{encoder-only}) we trained two triples of models $\left(\CtoA, \CAtoQ, \CQtoA\right)$ on the extractive subsets of SQuAD2 and NQ. We extracted 8M unlabeled windows of 512 tokens from the NQ training set. For each unlabeled window we generated  one example from the SQuAD2-trained models and one example from the NQ-trained models. For $A$ we picked an answer uniformly from the top 10 extractive answers according to $p(a | c; \CtoA)$. For $A'$ we picked the best extractive answer according to $p(a | c, q; \CQtoA)$. Filtering for roundtrip consistency gave us 2.4M and 3.2M synthetic positive instances from SQuAD2- and NQ-trained models respectively. We then added synthetic unanswerable instances by taking the question generated from a window and associating it with a non-overlapping window from the same Wikipedia page. We then sampled negatives to obtain a total of 3M and 4M synthetic training instances for SQuAD2 and NQ respectively. We trained models analogous to \newcite{alberti2019bert} initializing from the public BERT model, with a batch size of 128 examples for one epoch on each of the two sets of synthetic examples and on the union of the two, with a learning rate of $2 \cdot 10^{-5}$ and no learning rate decay. We then fine-tuned the the resulting models on SQuAD2 and NQ. 

\footnotetext{\url{https://github.com/google-research/bert}}

In our full pretraining experiments (Section \ref{seq2seq}) we only trained $\left(\CtoA, \CAtoQ, \CQtoA\right)$ on SQuAD2. However, we pretrained our question generation model on all of the BERT pretraining data, generating the next sentence left-to-right. We created a synthetic, roundtrip filtered corpus with 50M examples. We then fine-tuned the model on SQuAD2 as previously described. We experimented with both the single model setting and an ensemble of 6 models.

\subsection{Results}
The final results are shown in Tables \ref{tab:squad2_results} and \ref{tab:nq_results}. We found that pretraining on SQuAD2 and NQ synthetic data increases the performance of the fine-tuned model by a significant margin. On the NQ short answer task, the relative reduction in headroom is 50\% to the single human performance and 10\% to human ensemble performance. We additionally found that pretraining on the union of synthetic SQuAD2 and NQ data is very beneficial on the SQuAD2 task, but does not improve NQ results.

The full pretraining approach with ensembling obtains the highest EM and F1 listed in Table \ref{tab:squad2_results}. This result is only $0.1-0.4\%$ from human performance and is the third best model on the SQuAD2 leaderboard as of this writing (5/31/19).

\paragraph{Roundtrip Filtering} Roundtrip filtering appears to be consistently beneficial. As shown in Figure~\ref{fig:learning-curves}, models pretrained on roundtrip consistent data outperform their counterparts pretrained without filtering. From manual inspection, of 46 $(C, Q, A)$ triples that were roundtrip consistent 39\% were correct, while of 44 triples that were discarded only 16\% were correct.

\paragraph{Data Source} Generated question-answer pairs are illustrative of the differences in the style of questions between SQuAD2 and NQ. We show a few examples in Table~\ref{tab:samples}, where the same passage is used to create a SQuAD2-style and an NQ-style question-answer pair. The SQuAD2 models seem better at creating questions that directly query a specific property of an entity expressed in the text. The NQ models seem instead to attempt to create questions around popular themes, like famous works of art or TV shows, and then extract the answer by combining information from the entire passage.

\section{Conclusion}

We presented a novel method to generate synthetic QA instances and demonstrated improvements from this data on SQuAD2 and on NQ. We additionally proposed a possible direction for formal grounding of this method, which we hope to develop more thoroughly in future work.

\bibliography{acl2019}
\bibliographystyle{acl_natbib}

\clearpage
\newpage

\appendix

\section{Supplementary Material: a Sketch of a Formal Justification for the Approach}

\commentout{
A key question for future work is to develop a more formal understanding
of why the roundtrip method improves accuracy on question answering tasks
(similar questions arise for the back-translation methods of \cite{??}; 
a similar theory may apply to these methods). In this section we sketch
a possible approach, inspired by the method of \newcite{Balcan:2005} for 
learning with labeled and unlabeled data. This section is intentionally 
rather speculative but is intended to develop intuition about the methods, and
to propose possible directions for future work for a formal grounding.}

This section sketches a potential approach to giving a formal justification
for the roundtrip method, inspired by the method of \newcite{Balcan:2005} for 
learning with labeled and unlabeled data. This section is intentionally 
rather speculative but is intended to develop intuition about the methods, and
to propose possible directions for future work in developing a more formal grounding.

Assume that we have parameter estimates $\eCtoA$ and $\eCAtoQ$ derived from
labeled examples.
The log-likelihood function for the remaining parameters
is then
\begin{eqnarray*}
L(\CQtoA) &=& \sum_i \log p(a^{(i)} | q^{(i)}, c^{(i)}; \CQtoA)
\end{eqnarray*}
Estimation from labeled examples alone would involve the following
optimization problem:
\begin{equation}
{\eCQtoA} = \operatorname{argmax}_{\CQtoA \in {\cal H}} 
L(\CQtoA)
\label{eq:estimate}
\end{equation}
where ${\cal H}$ is a set of possible parameter values---typically ${\cal H}$
would be unconstrained, or would impose some regularization on 
$\CQtoA$.

Now assume we have some auxiliary function $\aux(\CQtoA)$ that measures
the roundtrip consistency of parameter values $(\CQtoA, \eCtoA, \eCAtoQ)$ on a set of
unlabeled examples. We will give concrete proposals for $\aux(\CQtoA)$ below.
A natural alternative to Eq.~\ref{eq:estimate} is then
to define
\[
{\cal H}' = \{ 
\CQtoA \in {\cal H}: 
\aux(\CQtoA) \geq \gamma
\}
\]
for some value of $\gamma$,
and to derive new parameter estimates
\begin{equation}
{\eCQtoA} = \operatorname{argmax}_{\CQtoA \in {\cal H}'} 
L(\CQtoA)
\label{eq:estimate2}
\end{equation}
The value for $\gamma$ can be estimated using cross-validation of accuracy on tuning data.

Intuitively a good choice of auxiliary function $\aux(\CQtoA)$ would have the 
property that there is some value of $\gamma$ such that: (1) ${\cal H}'$ is much "smaller"
or less complex than ${\cal H}$, and hence many fewer labeled examples are required 
for estimation (\newcite{Balcan:2005} give precise guarantees of this type); (2) ${\cal H}'$
nevertheless contains `good' parameter values that perform well on the labeled data.

A first suggested auxiliary function is the following, which makes use of unlabeled examples
$c^{(j)}$ for $j = 1 \ldots m$:
\begin{small}
\begin{eqnarray*}
\hspace{-0.5cm}&& \aux(\CQtoA) = \nonumber\\
&&\frac{1}{m}\sum_j \sum_{a, q} p(a, q | c^{(j)}; \eCtoA, \eCAtoQ) 
\log p(a | q, c^{(j)}; \CQtoA)
\end{eqnarray*}
\end{small}
where
$p(a, q | c^{(j)}; \eCtoA, \eCAtoQ) = p(a | c^{(j)}; \eCtoA) \times p(q | a, c^{(j)}; \eCAtoQ)$.

This auxiliary function encourages roundtrip consistency under parameters $\CQtoA$.
It is reasonable to assume that the optimal parameters achieve a high value for $\aux(\CQtoA)$, and hence that this will be a useful auxiliary function.

A second auxiliary function, which may be more closely related to the approach in the current
paper, is derived as follows. Assume we have some method of deriving triples $(c^{(j)}, q^{(j)}, a^{(j)})$
from unlabeled data, where a significant proportion of these examples are `correct' question-answer pairs. Define the following auxiliary function:
\begin{eqnarray*}
\hspace{-0.5cm}&& \aux(\CQtoA) = \nonumber\\
&&\frac{1}{m}\sum_j 
f(p(a^{(j)} | q^{(j)}, c^{(j)}; \CQtoA))
\end{eqnarray*}
 Here $f$ is some function that encourages high values for $p(a^{(j)} | q^{(j)}, c^{(j)}; \CQtoA)$. One choice would be $f(z) = \log z$; another choice would be $f(z) = 1$ if $z \geq \mu$, $0$ otherwise, where $\mu$ is a target `margin'. Thus under this auxiliary function the constraint
\[
\beta(\CQtoA) \geq \gamma
\]
would force the parameters $\CQtoA$ to fit the triples $(c^{(j)}, q^{(j)}, a^{(j)})$ derived from unlabeled data.

A remaining question is how to solve the optimization problem in Eq.~\ref{eq:estimate2}. One obvious approach would be to perform gradient ascent on the objective
\[
L(\CQtoA)+ \lambda \aux(\CQtoA)
\]
where $\lambda > 0$ dictates the relative weight of the two terms, and can be estimated using cross-validation on tuning data (each value for $\lambda$ implies a different value for $\gamma$). 

A second approach may be to first pre-train the parameters on the auxiliary function 
$\aux(\CQtoA)$, then fine-tune on the function $L(\CQtoA)$. In practice this may lead
to final parameter values with relatively high values for both objective functions. This latter approach appears to be related to the algorithms described in the current paper; future work should investigate this more closely.

\end{document}